\documentclass[runningheads]{llncs}
\usepackage[T1]{fontenc}
\usepackage{lmodern}

\usepackage{graphicx}
\usepackage{hyperref}

\usepackage[misc]{ifsym}

\usepackage{amsmath}
\usepackage{amsfonts}
\usepackage{multirow}
\usepackage{booktabs}
\usepackage{threeparttable}
\usepackage{setspace}
\usepackage{microtype}
\usepackage{xspace}
\usepackage{tablefootnote}

\usepackage{pifont}
\newcommand{\cmark}{\ding{51}}
\newcommand{\xmark}{\ding{55}}

\newcommand{\modelname}{$\mu\text{KG}$\xspace}
\newcommand{\repeatthanks}{\textsuperscript{\thefootnote}}

\begin{document}
\title{$\mu\text{KG}$: A Library for Multi-source Knowledge Graph Embeddings and Applications}
\titlerunning{$\mu\text{KG}$: A Library for Multi-source Knowledge Graph Embeddings}
%
\author{
    Xindi Luo\thanks{Equal contributors}\and
    Zequn Sun\repeatthanks\and
    Wei Hu\textsuperscript{(\Letter)}
}

\authorrunning{X. Luo et al.}
%
\institute{
	State Key Laboratory for Novel Software Technology,\\ 
	Nanjing University, Nanjing, China \\
	National Institute of Healthcare Data Science,\\
	Nanjing University, Nanjing, China \\
	\email{\{xdluo,zqsun\}.nju@gmail.com, whu@nju.edu.cn}
}
\maketitle              
\begin{abstract}
This paper presents $\mu\text{KG}$, an open-source Python library for representation learning over knowledge graphs.
$\mu\text{KG}$ supports joint representation learning over multi-source knowledge graphs (and also a single knowledge graph), 
multiple deep learning libraries (PyTorch and TensorFlow2), 
multiple embedding tasks (link prediction, entity alignment, entity typing, and multi-source link prediction), 
and multiple parallel computing modes (multi-process and multi-GPU computing).
It currently implements 26 popular knowledge graph embedding models and supports 16 benchmark datasets.
$\mu\text{KG}$ provides advanced implementations of embedding techniques with simplified pipelines of different tasks.  
It also comes with high-quality documentation for ease of use.
$\mu\text{KG}$ is more comprehensive than existing knowledge graph embedding libraries.
It is useful for a thorough comparison and analysis of various embedding models and tasks.
We show that the jointly learned embeddings can greatly help knowledge-powered downstream tasks, such as multi-hop knowledge graph question answering.
We will stay abreast of the latest developments in the related fields and incorporate them into $\mu\text{KG}$.

\medskip
\textbf{Resource Type}: Software

\smallskip
\textbf{License:} GPL-3.0 License

\smallskip
\textbf{GitHub Repository}: \url{https://github.com/nju-websoft/muKG}

\end{abstract}

\section{Introduction}
\label{sect:intro}
Knowledge graphs (KGs), such as Freebase~\cite{Freebase}, DBpedia~\cite{DBpedia}, Wikidata~\cite{Wikidata}, and YAGO~\cite{Yago}, store rich structured knowledge about the real world. 
They have been widely used in a variety of knowledge-driven applications, including semantic search, question answering, and logic reasoning~\cite{KG_Survey}.
Learning vector representations (a.k.a. embeddings) of KGs has become critical to support these intelligent applications.
In the past ten years, various KG embedding models such as TransE \cite{TransE}, ConvE \cite{ConvE}, RotatE \cite{RotatE} and TuckER \cite{TuckER} were proposed and achieved promising performance.
Please refer to the recent surveys \cite{LP_survey,KGE_survey} for an overview.
With these applications becoming more and more popular and diverse, they put forward higher demands to KGs in terms of coverage, richness and multilingualism. 
Oftentimes, a single KG cannot meet all these demands. 
This difficulty calls for the integration of multiple KGs. 
Learning from multi-source KGs with entity alignment has drawn a lot of attention in recent years \cite{MTransE,JAPE}.
The joint KG embeddings have demonstrated useful for a variety of downstream tasks such as entity typing and multi-source KG completion \cite{KGC_EA_EMNLP,KGC_EA_AKBC}.

\begin{table}
  \centering 
  \caption{Comparison of existing KG embedding libraries and ours.}
  \label{tab:summary}
  \resizebox{\textwidth}{!}{\setlength\tabcolsep{6pt}
  \begin{tabular}{lccccccc}
    \toprule
    \multirow{2}{*}{Libraries} & \multirow{2}{*}{Multi-KG support} & \multicolumn{2}{c}{Deep learning libraries} & \multicolumn{4}{c}{KG tasks} \\
    \cmidrule{3-8} 
    & & PyTorch & TensorFlow & LP & EA & ET & Multi-LP \\
    \midrule
    OpenKE~\cite{OpenKE}~\tablefootnote{\url{https://github.com/thunlp/OpenKE}} & \xmark & \cmark & TF1 & \cmark & \xmark & \xmark & \xmark \\
    DGL-KE~\cite{DGLKE}~\tablefootnote{\url{https://github.com/awslabs/dgl-ke}} & \xmark & \cmark & \xmark & \cmark & \xmark & \xmark & \xmark \\
    Pykg2vec~\cite{Pykg2vec}~\tablefootnote{\url{https://github.com/Sujit-O/pykg2vec}} & \xmark & \cmark & TF2 & \cmark & \xmark & \xmark & \xmark \\
    PyKEEN~\cite{KEEN,PyKEEN}~\tablefootnote{\url{https://github.com/pykeen/pykeen}} & \xmark & \cmark & \xmark & \cmark & \xmark & \xmark & \xmark \\
    TorchKGE~\cite{TorchKGE}~\tablefootnote{\url{https://github.com/torchkge-team/torchkge}} & \xmark & \cmark & \xmark & \cmark & \xmark & \xmark & \xmark \\
    LibKGE~\cite{LibKGE}~\tablefootnote{\url{https://github.com/uma-pi1/kge}} & \xmark & \cmark & \xmark & \cmark & \xmark & \xmark & \xmark \\ 
    OpenEA~\cite{OpenEA}~\tablefootnote{\url{https://github.com/nju-websoft/OpenEA}} & \cmark & \xmark & TF1 & \xmark & \cmark & \xmark & \xmark \\
    EAkit~\cite{EAkit}~\tablefootnote{\url{https://github.com/THU-KEG/EAkit}} & \cmark & \cmark & \xmark & \xmark & \cmark & \xmark & \xmark \\
    NeuralKG~\cite{NeuralKG}~\tablefootnote{\url{https://github.com/zjukg/NeuralKG}} & \xmark & \cmark & \xmark & \cmark & \xmark & \xmark & \xmark \\
    \midrule
    \modelname (ours) & \cmark & \cmark & TF2 & \cmark & \cmark & \cmark & \cmark \\
    \bottomrule
  \end{tabular}}
\end{table}

To support the easy use of KG embeddings and foster reproducible research into KG embedding techniques, much effort has been dedicated to developing KG embedding libraries, including OpenKE~\cite{OpenKE}, DGL-KE~\cite{DGLKE}, Pykg2vec~\cite{Pykg2vec}, PyKEEN~\cite{PyKEEN,KEEN}, TorchKGE~\cite{TorchKGE}, LibKGE~\cite{LibKGE}, OpenEA~\cite{OpenEA}, EAkit~\cite{EAkit} and NeuralKG \cite{NeuralKG}.
The majority of these libraries concentrates on the typical KG embedding task of link prediction.
Only OpenEA and EAkit support multi-source KG embedding and the corresponding task entity alignment.
Besides, most of them only support one deep learning library, especially PyTorch.
No one supports another prominent deep learning library TensorFlow2 (TF2 for short).
This limits the contexts in which these libraries can be used.
Facing these limitations of existing work and being aware of the effectiveness of multi-source KG embeddings, we develop a new scalable library, namely \modelname, for multi-source KG embeddings and applications.
Table~\ref{tab:summary} compares \modelname with the existing popular KG embedding libraries.
In summary, \modelname has the following features:
\begin{itemize}
\item\textbf{Comprehensive}. 
\modelname is a full-featured Python library for representation learning over a single KG or multi-source KGs. 
It is compatible with the two widely-used deep learning libraries PyTorch and TF2, and can therefore be easily integrated into downstream applications.
It integrates a variety of KG embedding models and supports four KG tasks including link prediction, entity alignment, entity typing, and multi-source link prediction.

\smallskip\item\textbf{Fast and scalable}. 
\modelname provides advanced implementations of KG embedding techniques with the support of multi-process and multi-GPU parallel computing, making it fast and scalable to large KGs.

\smallskip\item\textbf{Easy-to-use}. 
\modelname provides simplified pipelines of KG embedding tasks for easy use.  
Users can interact with \modelname through both methods APIs and command line.
It also has high-quality documentation.

\smallskip\item\textbf{Open-source and continuously updated}. 
The source code of \modelname is publicly available.
Our team will keep up-to-date on new related techniques and integrate new (multi-source) KG embedding models, tasks, and datasets into \modelname.
We will also keep improving existing implementations.

\end{itemize}

Our experiments on several benchmark datasets demonstrate the effectiveness and efficiency of our library \modelname.
Moreover, we carefully design two new tasks, multi-source link prediction and multi-source knowledge graph question answering (KGQA), with experiments to demonstrate the potential of multi-source KG embeddings:
\begin{itemize}

\item For \textbf{multi-source link prediction},
we can convert the multiple KGs into a joint graph by merging their aligned entities, on which we learn joint KG embeddings for link prediction over each KG.
This differs from the traditional link prediction, which first trains the model on a single KG and then predicts links for the same KG.
In our joint learning setting, to avoid the test set of a KG's link prediction task having overlap with other KGs' training set, 
we do not consider relation alignment in multi-source KGs, and also remove these overlapping triples from the training set if they exist.
Our experiment on DBP15K \cite{JAPE} shows that the joint trained TransE \cite{TransE} outperforms its separately trained variant by $122\%$ on $Hits@1$. 

\smallskip\item For \textbf{multi-source KGQA}, as a downstream application of KGs, we have attempted to use multi-source KG embeddings to aid in the task of multi-hop question answering over a KG. 
The typical pipeline of using KG embeddings to answer natural language questions \cite{EmbedKGQA} is learning to align the question representation (encoded by a pre-trained language model like BERT \cite{Bert}) with the answer entity's embedding (encoded by a KG embedding model like ComplEx \cite{ComplEx}).
Conventional methods and datasets only consider QA over a single KG.
We introduce an additional KG for joint embedding with the target KG using \modelname.
The embeddings of the target KG from the joint space are used for QA.
Our results on WebQuestionsSP \cite{WebQuestionsSP} show that joint KG embeddings can improve the accuracy by $8.6\%$ over independently trained embeddings on a single KG.
\end{itemize}

Overall, these experiments show that the multi-source KG embeddings are able to promote knowledge fusion and transfer, and therefore benefit downstream tasks.
We hope that our \modelname library can encourage the use of multi-source KG embeddings and promote their applications.

\section{\modelname}

\begin{figure}[!t]
\includegraphics[width=\textwidth]{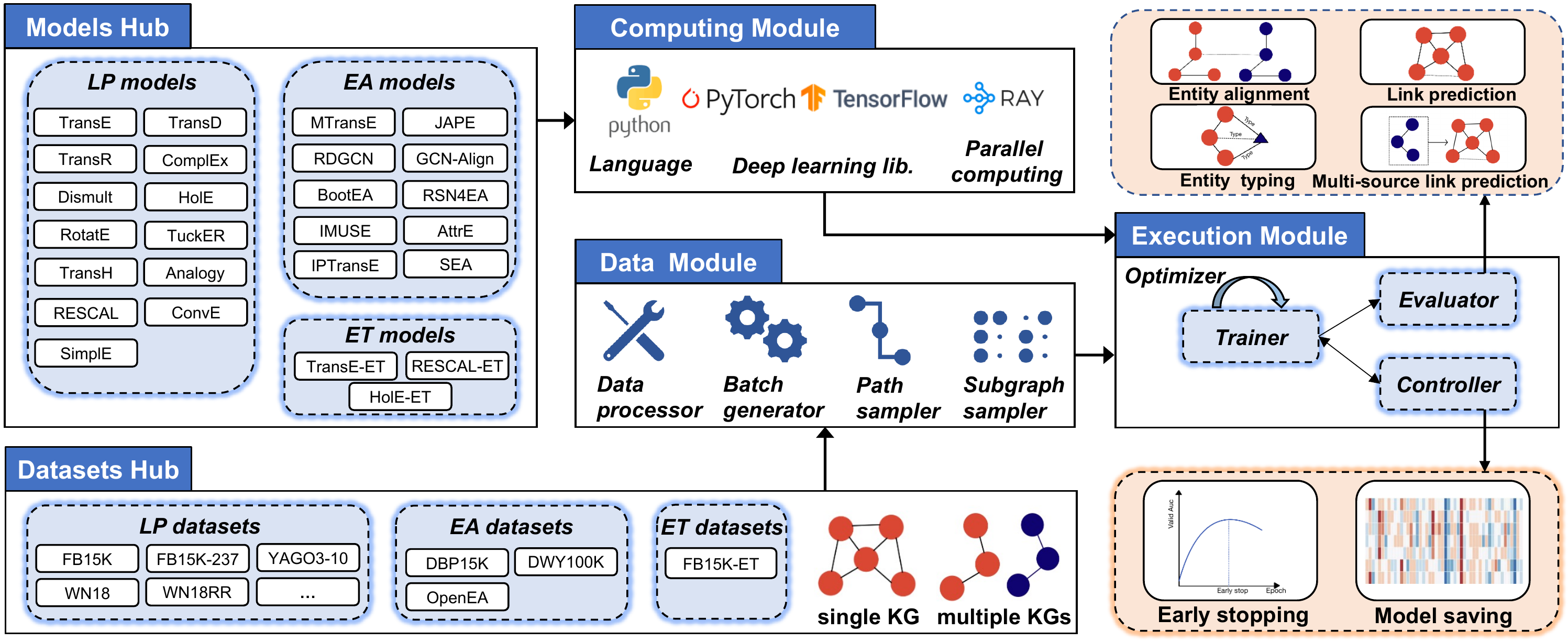}
\caption{Framework overview of \modelname.}
\label{fig:framework}
\centering
\end{figure}

\modelname is a scalable library for multi-source KG embeddings and applications.
It also supports representation learning over a single KG.
Its architecture is shown in Fig.~\ref{fig:framework}. 
\modelname supports a variety of link prediction, entity alignment, and entity typing models, as well as the datasets that go with them.
It consists of three modules.
The \textit{data module} converts the input single KG or multi-source KGs into the training data format (e.g., triples, paths or subgraphs) used by the embedding model.
The \textit{computing module} supports the execution module with neural computation and parallel training solutions.
As a result, the \textit{execution module} can be used for large-scale KGs and is compatible with the widely-used deep learning libraries PyTorch and TensorFlow.
The execution module trains a KG embedding model with the training data produced by the data module. 
The \textit{controller} keeps track of and records the training process.
The \textit{evaluator} employs the pre-trained embedding model to perform KG tasks, such as link prediction, entity alignment, entity typing, and multi-source link prediction.

\subsection{Data Module}
We hereby introduce the data module of \modelname.

\subsubsection{Data processor.}

The goal of the \textit{data processor} is to generate numerical IDs for entities, relations, and attributes from the input single KG or multi-source KGs in the Datasets Hub.
The numerical ID is the identifier of a resource in an embedding model.
The data processor first reads the original relation triples and attribute triples from the \texttt{txt} or \texttt{ttl} files. 
Then, it assigns an ID to each entity, relation, and attribute.
It currently provides two ID generation algorithms.
The \textit{unique-ID algorithm} generates a unique ID for each resource in KGs.
It can be used for both single KG and multi-source KGs.
The \textit{shared-ID algorithm} generates the same ID for aligned entities in different KGs. 
In this way, the multiple KGs are merged as a ``single'' joint graph. 

\subsubsection{Batch generator.}

The \textit{batch generator} takes as input KG triples and divides the complete data into multiple fixed-size batches for model training.
If the model requires relational paths or subgraphs for training,
the batch generator would first call the path or subgraph sampler to convert triples.
The batch generator also includes several negative sampling methods to randomly generate negative examples (e.g., negative alignment pairs or negative triples) for each positive example.
The positive and negative examples are used in the embedding model for contrastive embedding learning. 
The \textit{uniform negative sampling} method replaces an entity in a triple or an alignment pair with another randomly-sampled entity to generate a negative example.
It gives each entity the same replacement probability. 
Such uniform negative sampling has the problem of inefficiency since many sampled negative samples are obviously false as training goes on, which does not provide any meaningful information.
\modelname also supplies the self-adversarial negative sampling method \cite{RotatE} and the truncated negative sampling method \cite{BootEA} that seek to generate hard negative examples.

\subsubsection{Path sampler.}
The \textit{path sampler} is to support some embedding models that are built by modeling the paths of KGs, such as IPTransE \cite{IPTransE} and RSN \cite{RSN}.
It can generate three types of paths based on random walks.
The first is the relational path like $(e_1, r_1, e_2, r_2, e_3)$,
where $e_i$ stands for an entity and $r_j$ denotes a relation.
It is an entity-relation chain.
The second is the entity path like $(e_1, e_2, e_3)$, and the third is the relation path like $(r_1, r_2)$.

\subsubsection{Subgraph sampler.}
The \textit{subgraph sampler} is to support GNN-based embedding models like GCN-Align \cite{GCN-Align} and AliNet \cite{AliNet}.
It can generate both first-order (i.e., one-hop) and high-order (i.e., multi-hop) neighborhood subgraphs of entities.
The GNN-based models represent an entity by aggregating the embeddings of its neighbors in the subgraphs.

\subsection{Execution Module}
This module carries out the training task of embedding models.

\subsubsection{Trainer.}

The \textit{trainer} directs the model training and evaluation based on the detailed configurations of users.
It manages the model's training progress.
\modelname configures trainers for entity alignment models, link prediction models, and entity typing models, respectively.
The \textit{trainer} provides three optimizers, including the standard stochastic gradient descent, Adagrad, and Adam.
It implements four loss functions, including the mean-squared loss,
marginal ranking loss, limit-based loss, and noise-contrastive estimation loss.

\subsubsection{Evaluator.}

The \textit{evaluator} is to assess the performance of the trained model on specific test data.
For (joint) link prediction, it uses the energy function to compute the plausibility of a candidate triple.
For entity alignment or typing, it provides several metrics to measure entity embedding similarities, such as the cosine, inner, Euclidean distance, and cross-domain similarity local scaling.
The evaluation process can be accelerated using multi-processing.
The implemented metrics for assessing the performance of embedding tasks include $Hits@K$, mean rank ($MR$) and mean reciprocal rank ($MRR$).     
$Hits@K$ measures the percentage of the test cases in which the correct counterpart is ranked in the top $k$.
$MR$ calculates the mean of these ranks. 
$MRR$ is the average of the reciprocal ranks of results. 
Higher $Hits@K$ and $MRR$ or lower $MR$ values indicate better performance.

\subsubsection{Controller.}

The \textit{controller} is in charge of the trainer.
During the training process, the controller calls the evaluator to assess the model performance on validation data.
If the performance begins to drop continuously,
the controller would terminate the training (i.e., early stopping).
After that, the controller saves the model and embeddings for further use.

\subsection{Computing Module}
In this section, we introduce the computing module.

\subsubsection{Support of PyTorch and TF2.}
The computing module uses PyTorch and TF2 as the backbone for neural computing.
Users can choose one of the backbones to run \modelname or carry on secondary development based on \modelname.
If no backbone is specified by the user,
\modelname can automatically detect which backbone has already been installed in the Python environment.

\subsubsection{Multi-GPU and multi-processing computation.}
\begin{figure}[!t]
\centering
\includegraphics[width=\textwidth]{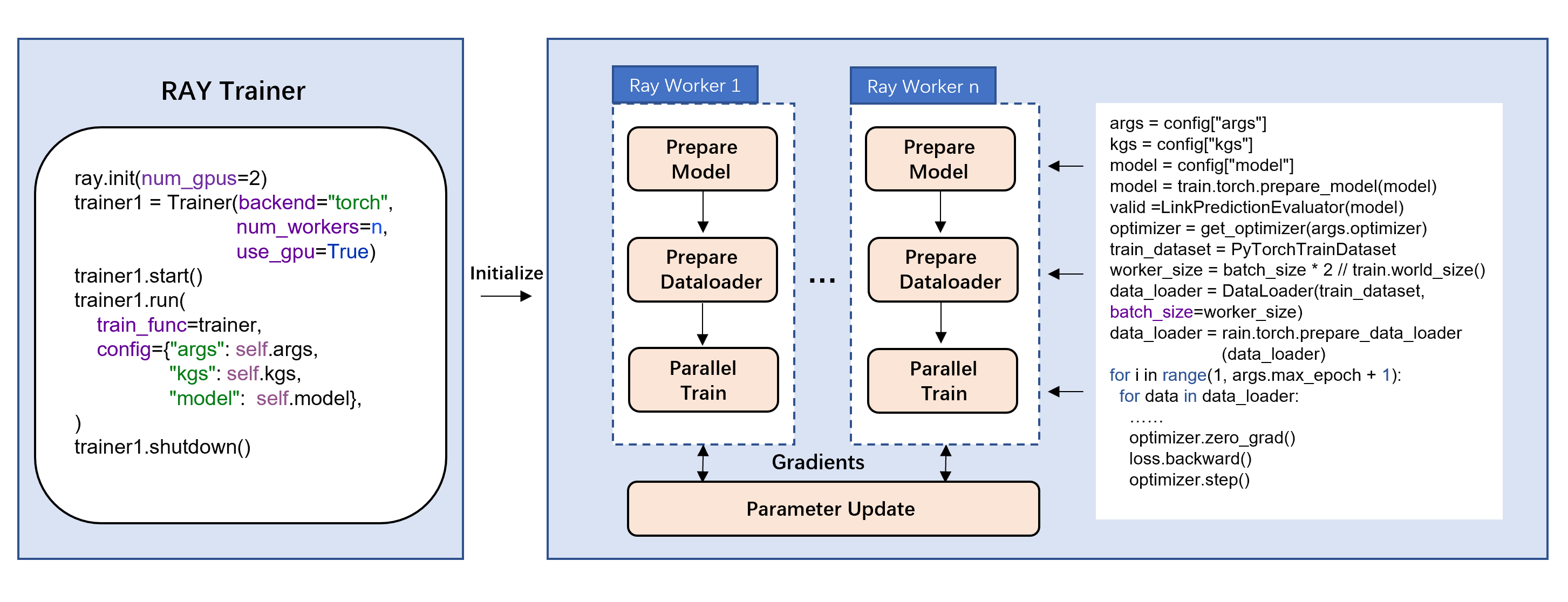}
\caption{Code snippet for training KG embedding models in the parallel mode.}
\label{fig:ray}
\end{figure}

Scalability is a key consideration when we develop \modelname, because KGs in real-world applications are typically very large.
Although PyTorch and TensorFlow both provide interfaces for parallel computing,
they differ greatly in implementation and are difficult for users to use.
Hence, we use Ray\footnote{\url{https://www.ray.io/}} to provide a uniform and easy-to-use interface for multi-GPU and multi-processing computation.
Fig.~\ref{fig:ray} shows our Ray-based implementation for parallel computing and the code snippet to use it.
Users can set the number of CPUs or GPUs used for model training.

\subsection{User Interface}

\begin{figure}[!t]
\centering
\includegraphics[width=\textwidth]{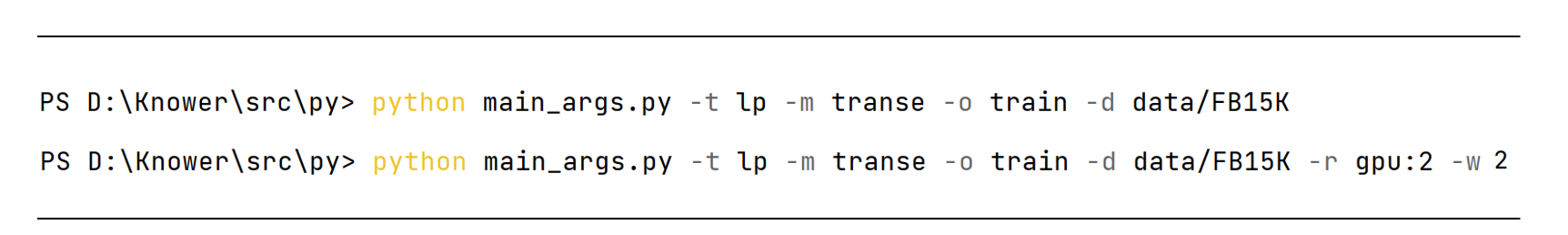}
\caption{Command line for using \modelname.}
\label{fig:interface}
\end{figure}

\modelname gives users two options for running KG embedding models.
For users that are unfamiliar with \modelname,
they can run a model on a dataset with the command line, as shown in Fig.~\ref{fig:interface}.
For advanced users, they can modify the configurations of a model and call the model's running function in their Python code.

\section{Experiments}\label{sect:exp}
In this section, we report our experiments to evaluate the effectiveness and efficiency of \modelname.
The source code is available at our GitHub repository.\footnote{\url{https://github.com/nju-websoft/muKG}}

\subsection{Experiments on Effectiveness}
To evaluate the effectiveness, we compare the results produced by our library with the corresponding official results reported in the models' papers.
We consider one single-KG task \textit{link prediction}, and three multi-KG tasks \textit{entity alignment}, \textit{entity typing} and \textit{multi-source link prediction}.

\begin{table}[!t]
\centering
\makebox[0pt][c]{\parbox{\textwidth}{%
    \begin{minipage}[b]{0.48\hsize}\centering
        \caption{LP results on FB15K.}
        \label{tab:fb15k}
        \resizebox{\textwidth}{!}{\setlength\tabcolsep{2pt}
        \begin{tabular}{lcccc}
		    \toprule
			Models & & $Hits@1$ & $Hits@10$ & $MRR$ \\ 
		 	\midrule
		 	\multirow{2}{*}{RESCAL\xspace} & original & $-$& 0.284& $-$   \\
			\cmidrule{2-5}
		    & ours & 0.129 & 0.342 & 0.202 \\
		    \midrule
			\multirow{2}{*}{TransE} & original & $-$ & 0.471 & $-$ \\
			\cmidrule{2-5}
			& ours & 0.194 & 0.647 & 0.353\\
			\midrule
			\multirow{2}{*}{TransH} & original & $-$& 0.585& $-$ \\
			\cmidrule{2-5}
			& ours & 0.188 & 0.604 & 0.332 \\
			\midrule
			\multirow{2}{*}{TransD} & original & $-$& 0.742& $-$  \\
			\cmidrule{2-5}
			& ours & 0.214 & 0.595 & 0.345 \\
			
			\bottomrule
        \end{tabular}}
    \end{minipage}
    \hfill
    \begin{minipage}[b]{0.47\hsize}\centering
        \caption{LP results on FB15K-237.}
        \label{tab:fb15k237}
        \resizebox{\textwidth}{!}{\setlength\tabcolsep{2pt}
        \begin{tabular}{lcccc}
		    \toprule
			Models & & $Hits@1$ & $Hits@10$ & $MRR$ \\ 
		 	\midrule
			\multirow{2}{*}{TransE} & original & $-$ & 0.465 & 0.294 \\
			\cmidrule{2-5}
			& ours & 0.174& 0.463 & 0.270 \\
			\midrule
			\multirow{2}{*}{ConvE} & original &0.237 & 0.501 &0.325 \\
			\cmidrule{2-5}
			& ours & 0.237 & 0.514 & 0.327\\
			\midrule
			\multirow{2}{*}{RotatE} & original & 0.205 & 0.480 & 0.297\\
			\cmidrule{2-5}
			& ours &0.172 & 0.456 & 0.260 \\
			\midrule
			\multirow{2}{*}{TuckER\xspace} & original & 0.266 & 0.544 & 0.358\\
			\cmidrule{2-5}
			& ours & 0.254 & 0.535 & 0.346\\
			\bottomrule
        \end{tabular}}
    \end{minipage}
}}
\end{table}

\subsubsection{Link prediction.}
We choose two benchmark datasets, FB15K \cite{TransE} and FB15K-237 \cite{FB15K237}, for link prediction evaluation.
FB15K-237 was created from FB15K to ensure that the testing and evaluation datasets do not have inverse relation test leakage. 
Recent link prediction models use FB15K-237 for evaluation.
On FB15K, we compare the $Hits@1$, $Hits@10$ and $MRR$ results of four old but popular models in Table~\ref{tab:fb15k}, including RESCAL \cite{RESCAL}, TransE \cite{TransE}, TransH \cite{TransH}, and TransD \cite{TransD}.
``$-$'' denotes the unreported results.
We can see that our implemented RESCAL, TransE and TransH can achieve better results than the original code due to our modern implementations.
We also notice that the implemented TransD shows lower Hits@10 performance than its original version.
The reason lies in the different evaluation settings. 
The original TransD removes the corrupted triplets in the training, validation and test sets before ranking. 
But our implementation only removes those in the training set following other methods because this is more reasonable.
On FB15K-237, we compare TransE and other three recent models including ConvE \cite{ConvE}, RotatE \cite{RotatE},\footnote{We use uniform negative sampling for a fair comparison with other models.} and TuckER \cite{TuckER} in Table~\ref{tab:fb15k237}.
The results of TransE on FB15K-237 are taken from \cite{RotatE} because TransE was not evaluated on this dataset.
As we can see, our implementations of TransE and ConvE in \modelname perform very similarly to their original code.
As for RotatE and TuckER, the performance of our implementations is slightly lower than the original results, 
but also in the range of acceptance.
This is due to different hyperparameter settings. 
In consideration of GPU resources, we do not set the embedding dimension to 1,000, which is used in their original papers but would cost too much GPU memory. 
Generally, a large dimension leads to good performance.
In summary, our implementations of link prediction models can basically reproduce the reported results.

\subsubsection{Entity alignment.}
We use the recent benchmark dataset OpenEA \cite{OpenEA} for entity alignment evaluation.
OpenEA also provides the implementations of several entity alignment models using TensorFlow 1.12.
We choose three structure-based entity alignment models GCN-Align \cite{GCN-Align}, SEA \cite{SEA} and BootEA \cite{BootEA}, as well as an attribute-enhanced model IMUSE \cite{IMUSE}, as baselines.
We compare their $Hits@1$, $Hits@10$ and $MRR$ results on OpenEA's EN-DE and EN-FR 15K settings with our PyTorch-based implementations and TF2-based implementations in Table~\ref{tab:entity_align}.
We can see that the two implementations of a model in \modelname achieve similar performance.
For SEA and IMUSE, PyTorch-based implementations perform better than TF2-based implementations.
We think this is caused by the difference between the two backbones.
When compared to the results of OpenEA, \modelname achieves comparable results.
This demonstrates the efficacy of our implementations for entity alignment models.

\begin{table}[!t]
	\centering
	\caption{Entity alignment results on EN-DE and EN-FR 15K.}
	\label{tab:entity_align}
	\resizebox{\textwidth}{!}{\setlength\tabcolsep{6pt}
		\begin{tabular}{llcccccc}
			\toprule
			\multirow{2}{*}{Models} & \multirow{2}{*}{Backends} & \multicolumn{3}{c}{EN-DE} & \multicolumn{3}{c}{EN-FR}  \\
			\cmidrule(lr){3-5} \cmidrule(lr){6-8}
			& & $Hits@1$ & $Hits@10$ & $MRR$ & $Hits@1$ & $Hits@10$ & $MRR$   \\ 
			\midrule
			\multirow{3}{*}{GCN-Align} 
			& OpenEA & 0.481 & 0.753 & 0.571 & 0.338 & 0.680 & 0.451\\
			\cmidrule{2-8}
			& TF2 (ours) & 0.480 & 0.754 & 0.571& 0.335 & 0.670 & 0.446 \\
			\cmidrule{2-8}
			& PyTorch (ours) & 0.460 & 0.747 & 0.560 & 0.337 & 0.671 & 0.453 \\
			\midrule
			\multirow{3}{*}{SEA} 
			& OpenEA & 0.530 & 0.796 & 0.617 & 0.280 & 0.642 & 0.328\\
			\cmidrule{2-8}
			& TF2 (ours) & 0.536 & 0.806 & 0.624 & 0.281 & 0.630 & 0.304\\
			\cmidrule{2-8}
			& PyTorch (ours) & 0.561 & 0.834 & 0.650 & 0.321 & 0.679 & 0.439\\
			\midrule
			\multirow{3}{*}{BootEA} 
			& OpenEA  & 0.675 & 0.865 & 0.740 & 0.507 & 0.794 & 0.603\\
			\cmidrule{2-8}
			& TF2 (ours) & 0.671 & 0.866 & 0.737 & 0.503 & 0.786 & 0.597 \\
			\cmidrule{2-8}
			& PyTorch (ours) & 0.662 & 0.884 & 0.738 & 0.493 & 0.811 & 0.599\\
			\midrule
			 \multirow{3}{*}{IMUSE} 
			 & OpenEA & 0.580 & 0.778 & 0.647 & 0.569 & 0.777 & 0.638 \\
			\cmidrule{2-8}
			 & TF2 (ours) & 0.567 & 0.672 & 0.636 & 0.571 & 0.777 & 0.640 \\
			\cmidrule{2-8}
			& PyTorch (ours) & 0.596 & 0.804 & 0.670 & 0.564 & 0.776 & 0.640 \\
			\bottomrule
	\end{tabular}}
\end{table}

\subsubsection{Entity typing.}
Entity typing can be seen as a special link prediction task across an instance KG and an ontological KG.
For example, given (``Michael Jackson'', ``rdf:type'', \_), the task is to predict the target type ``/music/artist''.
We use the FB15K-ET dataset for evaluation \cite{FB15K-ET}.
FB15K-ET is an expansion of FB15K with entity types.
We follow \cite{FB15K-ET} to implement two baselines, RESCAL-ET and HolE-ET, for entity typing.
The two models are built based on the link prediction models RESCAL \cite{RESCAL} and HolE \cite{HolE}, respectively.
We compare our results with those in \cite{FB15K-ET} in Table~\ref{tab:entity_type}.
We can observe that our implementations achieve similar or even better performance than those in \cite{FB15K-ET},
demonstrating the effectiveness of \modelname in entity typing.

\begin{table}[!t]
	\centering
	\caption{Entity typing results on FB15K-ET.}
	\label{tab:entity_type}
	\resizebox{0.65\textwidth}{!}{\setlength\tabcolsep{6pt}
		\begin{tabular}{llccc}
          \toprule
          Models & & $Hits@1$ & $Hits@10$ &  $MRR$ \\
          \midrule
          \multirow{2}{*}{RESCAL-ET} & original & 0.097 & 0.376 & 0.190 \\
          \cmidrule{2-5}
          & ours  & 0.128 & 0.456 & 0.236\\
          \midrule
          \multirow{2}{*}{HolE-ET} & original & 0.133 & 0.382 & 0.220 \\
          \cmidrule{2-5}
          & ours  & 0.129 & 0.522 & 0.252\\
          \bottomrule
	\end{tabular}}
\end{table}

\subsubsection{Multi-source link prediction.}
This is a new task that we propose, which is inspired by both link prediction in a single KG and entity alignment between two KGs.
We believe that training embeddings solely on a KG for link prediction is ineffective because the KG may be very incomplete.
We introduce another background KG with entity alignment to the target KG for joint KG embedding learning.
We use the shared-ID generation method in \modelname to merge the two KGs and learn embeddings of the joint KG with a KG embedding model such as TransE \cite{TransE}.
When the learning progress is completed,
only the embeddings of the target KG are used to participate in link prediction.
For evaluation, we choose DBP15K\textsubscript{ZH-EN} \cite{JAPE}.
It is an entity alignment dataset,
and we denote the two KGs in DBP\textsubscript{ZH-EN} by DBP\textsubscript{ZH} and DBP\textsubscript{EN}, respectively.
Following TransE~\cite{TransE}, 
we divide triples into training, validation and test sets. 
Specifically, DBP\textsubscript{ZH} has $ 63,372 $ training triples, $ 3,522 $ validation triples and $ 3,520 $ test triples, while DBP\textsubscript{EN} has $ 85,627 $, $ 4,758 $ and $ 4,757 $, respectively.
Conventional link prediction is usually carried out on a single KG. However, 
for multi-source link prediction with entity alignment, 
it would be interesting to see the performance of link prediction based on the jointly-trained KG embeddings. 
Based on \modelname, we train a TransE model over the joint graph of DBP\textsubscript{ZH} and DBP\textsubscript{EN}.
We choose three translational models TransE \cite{TransE}, TransH \cite{TransH} and TransD \cite{TransD}; 
four semantic matching models DistMult \cite{DistMult}, HolE \cite{HolE}, ComplEx \cite{ComplEx} and Analogy \cite{Analogy}; as well as two neural models ProjE \cite{ProjE} and ConvE \cite{ConvE}, as baselines.
From Table~\ref{tab:joint_link}, we can see that \modelname (TransE) outperforms the translational and semantic matching models. 
ConvE achieves better results than our method, 
but its model complexity is also much higher than TransE. 
By encoding alignment information, \modelname (TransE) greatly outperforms TransE. 
The results demonstrate the joint training is effective to improve the separately-trained models on link prediction. 
We think that this is because the alignment information between two KGs can complement the incomplete relational structures of each other.

\begin{table}[!t]
	\centering
	\caption{Link prediction results with joint KG embeddings.}
	\label{tab:joint_link}
	\resizebox{0.9\textwidth}{!}{\setlength\tabcolsep{6pt}
		\begin{tabular}{lcccccc}
			\toprule
			\multirow{2}{*}{Models} & \multicolumn{3}{c}{DBP\textsubscript{ZH}} & \multicolumn{3}{c}{DBP\textsubscript{EN}}  \\
			\cmidrule(lr){2-4} \cmidrule(lr){5-7}
			& $Hits@1$ & $Hits@10$ & $MRR$ & $Hits@1$ & $Hits@10$ & $MRR$   \\ 
			\toprule
			TransE & 0.100 & 0.529 & 0.248 & 0.099 & 0.512 & 0.241 \\
			TransH & 0.103 & 0.519 & 0.274 & 0.125 & 0.535 & 0.263 \\
			TransD & 0.097 & 0.506 & 0.237 & 0.114 & 0.517 & 0.251 \\
			\toprule
			DistMult & 0.095 & 0.375 & 0.188 & 0.100 & 0.385 & 0.195 \\
			HolE & 0.114 & 0.327 & 0.186 & 0.122 & 0.405 & 0.221 \\
			ComplEx & 0.174 & 0.374 & 0.245 & 0.195 & 0.435 & 0.279 \\
			Analogy & 0.145 & 0.363 & 0.220 & 0.169 & 0.375 & 0.241 \\
			\toprule
			ProjE & 0.257 & 0.613 & 0.317 & 0.265 & 0.629 & 0.323 \\
			ConvE & 0.291 & 0.597 & 0.398 & 0.322 & 0.631 & 0.429 \\
			\toprule
			\modelname (TransE) & 0.222  & 0.549 & 0.331 & 0.252  & 0.585 & 0.363 \\
			\toprule
	\end{tabular}}
\end{table}

\subsection{Experiments on Efficiency}
In this section, we evaluate the efficiency of the proposed library \modelname.
The experiments were conducted on a server with an Intel Xeon Gold 6240 2.6GHz CPU, 512GB of memory and four NVIDIA Tesla V100 GPUs.

\subsubsection{Efficiency of multi-GPU training.}
\begin{figure}
\centering
\includegraphics[width=0.8\textwidth]{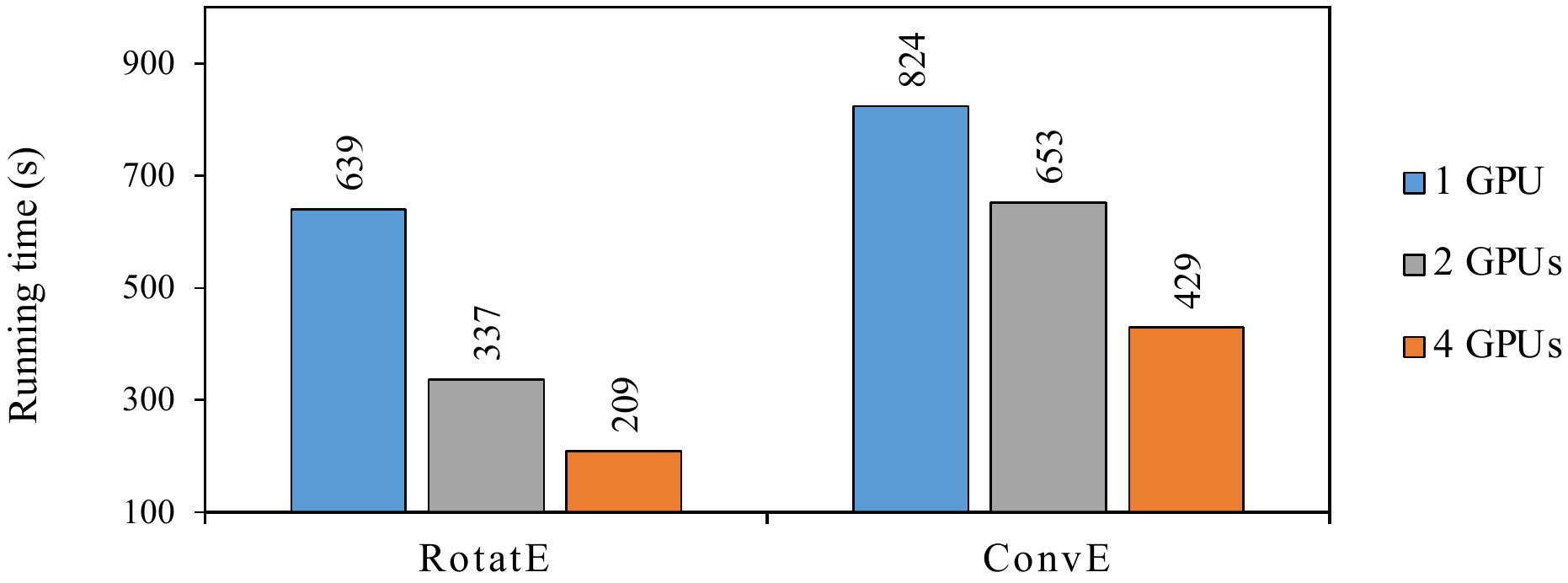}
\caption{Running time comparison on FB15K-237 with multi-GPU training.}
\label{fig:time}
\end{figure}

Fig.~\ref{fig:time} compares the training time of RotatE and ConvE on FB15K-237 when using different numbers of GPUs.
As we can see, using multiple GPUs for parallel computing can significantly accelerate training.
The final link prediction results are not affected by parallel computing.
For example, the $Hits@1$ scores of ConvE when using 1, 2, and 4 GPUs for computing are 0.241, 0.239 and 0.227, respectively.
This experiment shows the efficiency of our multi-GPU training.

\subsubsection{Efficiency comparison against LibKGE and PyKEEN.}
We further compare the training time used by \modelname with LibKGE \cite{LibKGE} and PyKEEN \cite{PyKEEN}. They are both PyTorch-based libraries for efficient training, evaluation, and optimization of KG embeddings.
The backbone of \modelname in this experiment is also PyTorch.
Table~\ref{tab:time_libkge} gives the training time of ConvE and RotatE on FB15K-237 with a single GPU for calculation.
For a fair comparison, we use the same hyper-parameter settings (e.g., batch size and maximum training epochs) for each model in the three libraries.
We discover that \modelname costs less time than LibKGE and PyKEEN to train a KG embedding model,
which demonstrates its efficiency.

\begin{table}
	\centering
	\caption{Running time on FB15K-237 with a GPU.}
	\label{tab:time_libkge}
	\resizebox{0.5\textwidth}{!}{\setlength\tabcolsep{6pt}
		\begin{tabular}{lccc}
          \toprule
          Models & \modelname & LibKGE & PyKEEN\\
          \midrule
          RotatE & \textbf{639} s & 3,260 s & 1,085 s \\
          ConvE & \textbf{824} s & 1,801 s & \ \ 961 s\\
          \bottomrule
	\end{tabular}}
\end{table}

\section{Application to Multi-hop KGQA}
We hereby report the experimental results on the downstream task, i.e., multi-hop KGQA, using our proposed joint embeddings.

\subsubsection{Settings.}
We follow EmbedKGQA \cite{EmbedKGQA}, a recent popular embedding-based KGQA method, to build a QA pipeline with our \modelname.
EmbedKGQA consists of three modules.
The KG embedding module learns embeddings for the input KG.
Existing KG embedding models such as TransE \cite{TransE} and ComplEx~\cite{ComplEx} can be chosen.
The question embedding module encodes natural language questions with the help of the pre-trained language model RoBERTa \cite{RoBERTa},
which is a new training recipe that improves on BERT and is widely used for encoding natural language text.
The answer selection module chooses the final answer based on the question and relation similarity scores.
Using KG embeddings to answer natural language questions can make it more effective in handling the relational sparsity in KGs.
The KG embedding model used in EmbedKGQA is ComplEx.
In our pipeline for QA, 
we use \modelname (the embedding model is also ComplEx) to learn joint embeddings based on the target KG and another background KG Wikidata5M \cite{Wikidata5M}, which is a subset of Wikidata with million-scale entities.
For a fair comparison, we keep other modules in our pipeline the same as those in EmbedKGQA.

\subsubsection{Dataset.}
We choose the popular multi-hop QA benchmark WebQuestionsSP \cite{WebQuestionsSP} as the dataset.
There are $4,737$ questions in total.
This dataset contains 1-hop and 2-hop questions that may be answered using Freebase entities \cite{Freebase}.
Following EmbedKGQA, we limit the KG to a subset of Freebase that contains all relational triples within 2-hops of any entity specified in the WebQuestionsSP questions.
We refine it further to include only those relations that are stated in the dataset.
There are a total of $1.8$ million entities and $5.7$ million triples in this selected KG (denoted as FB4QA in this paper) to support these questions.
The number of entity links between Wikidata5M and FB4QA is $493,987$.

\begin{table}[!t]\setlength\tabcolsep{6pt}
	\centering
	\caption{QA accuracy on WebQuestionsSP.}
	\label{tab:qa} 
	\resizebox{\textwidth}{!}{
		\begin{tabular}{lcc}
			\toprule
			& EmbedKGQA~\cite{EmbedKGQA} & EmbedKGQA + Wikidata5M \\
			\midrule
			Half-FB4QA & 0.485 & 0.547 \\
			Full-FB4QA & 0.587 & 0.646 \\
			Full-FB4QA w/ rel. pruning & 0.666 & 0.723 \\
			\bottomrule
	\end{tabular}}
\end{table}

\subsubsection{Results.}
Table~\ref{tab:qa} presents the QA accuracy.
To study the effect of KG sparsity on QA performance, 
following EmbedKGQA, the FB4QA is used for two settings: Half-FB4QA and Full-FB4QA.
The former randomly drops half of the triples in FB4QA to simulate an incomplete KG.
The latter uses the full FB4QA to learn entity embeddings.
Besides, in the Full-FB4QA w/ rel. pruning setting, 
a relation pruning strategy is employed to reduce the candidate answer space by filtering out the dissimilar relations with the key entity in the question.
We can see from the table that EmbedKGQA + Wikidata outperforms the baseline EmbedKGQA in all three settings.
This is because our learned embeddings of FB4QA can benefit from the background KG, and thus are more expressive than those in EmbedKGQA.
Both our method and EmbedKGQA in the Full-KG setting achieve better accuracy than the corresponding result in the Half-KG setting.
This demonstrates that KG incompleteness degrades the quality of KG embeddings,
and thus causes a decrease in performance.
Our method can improve the incompleteness issue in KGs through knowledge transfer from other background KGs.
We also consider extending LibKGE \cite{LibKGE} for this new task. 
We merge two KGs into a large graph and use LibKGE to learn KG embeddings. The accuracy is 0.718 in the setting of Full-FB4QA w/ rel. pruning, a similar performance compared with our \modelname. 
This result further shows the effectiveness of our library and the potential of multi-source KG embeddings.
In summary, this experiment demonstrates that multi-source KG embeddings are also effective in improving KG-related downstream tasks, 
and knowledge transfer between multi-source KGs is an alternative for boosting performance in real-world applications.

\section{Related Work}\label{sect:related_work}
In this section, we review the related work on KG embedding models and tasks, as well as existing libraries for KG embedding.

\subsection{Knowledge Graph Embedding Tasks and Models}
\subsubsection{Link prediction.}
TransE~\cite{TransE} introduces translational KG embeddings. It defines the score function $f_{\text{TransE}}(\tau) = ||\mathbf{h} + \mathbf{r} - \mathbf{t}||$\footnote{Hereafter, $ ||\cdot|| $ denotes the $ L_2 $ vector norm.} to measure the plausibility of relational triple $\tau=(h,r,t)$, where $h$, $r$ and $t$ denote the head entity, relation and tail entity, respectively.
Boldfaced letters denote the corresponding vector representations.
Although TransE performs well for modeling one-to-one relations, it encounters issues when dealing with more complex relations. For example, if $ (h, r, t_1) $ and $ (h, r, t_2) $ hold for a one-to-many relation $ r $, we have $ \mathbf{h} + \mathbf{r} \approx \mathbf{t}_1 $ and $ \mathbf{h} + \mathbf{r} \approx \mathbf{t}_2 $, then $ \mathbf{t}_1 \approx \mathbf{t}_2 $. If $ (h, r_1, t) $ and $ (h, r_2, t) $ hold for $ h $ and $ t $, we have $ \mathbf{r}_1 \approx \mathbf{r}_2 $.
To resolve these problems, several improved translational models, such as TransH~\cite{TransH}, TransR~\cite{TransR}, and TransD~\cite{TransD}, have been proposed. 
They enable entities to have relation-specific embeddings.
For example, TransH interprets a relation as a translation vector on a hyperplane, while TransR and TransD embed entities and relations in distinct vector spaces.
RotatE~\cite{RotatE} is an improved variant in the complex vector space.
Besides, semantic matching-based models exploit similarity-based functions to score relational triples. 
The scores are computed using bilinear functions in RESCAL~\cite{RESCAL}, DistMult~\cite{DistMult}, ComplEx~\cite{ComplEx} and SimplE~\cite{SimplE}, while HolE~\cite{HolE} replaces dot product with circular correlation. 
Embeddings are given analogical qualities in Analogy~\cite{Analogy}.
Recently, neural network-based models, including ProjE~\cite{ProjE}, ConvE~\cite{ConvE}, R-GCN~\cite{R-GCN}, ConvKB~\cite{ConvKB}, KBGAN~\cite{KBGAN} and LinkNBed~\cite{LinkNBed}, achieve superior link prediction performance.
\modelname currently supports TransE, TransR, TransH, TransD, TuckER, DisMult, ComplEx, HolE, Analogy, RESCAL, RotatE, SimplE and ConvE.

\subsubsection{Entity alignment.}
Embedding-based entity alignment models usually consist of two modules, i.e., KG embedding and alignment learning. 
For KG embedding based on relational facts, many models including MTransE \cite{MTransE}, IPTransE \cite{IPTransE}, JAPE \cite{JAPE}, KDCoE \cite{KDCoE}, BootEA \cite{BootEA}, SEA \cite{SEA}, AttrE \cite{AttrE}, MultiKE \cite{MultiKE} and TransEdge \cite{TransEdge} adopt TransE \cite{TransE} or its improved variants. 
Most of other models like GCN-Align \cite{GCN-Align}, RDGCN~\cite{RDGCN} and AliNet \cite{AliNet} adopt GCN due to its powerful representation learning ability. 
Other models like RSN \cite{RSN} use recurrent neural networks for KG embedding, respectively. 
In addition to relational facts, some models such as KDCoE, AttrE, MultiKE, RDGCN and IMUSE ~\cite{IMUSE} also exploit entity attributes for KG embedding and achieve good results. 
For alignment learning, IPTransE and KDCoE use the pair loss. 
Besides, JAPE, BootEA, AttrE, RSN and MultiKE let aligned entities in seed alignment share the same or similar embeddings by some tailored data processing skills, which can be also regarded as a special case of the pair loss. GCN-Align and RDGCN use the marginal ranking loss and AliNet uses the limit-based loss. 
To achieve better performance, some models including IPTransE, BootEA, KDCoE and TransEdge further employ semi-supervised learning. 
\modelname currently supports MTransE, AttrE, SEA, GCN-Align, RDGCN, IPTransE, JAPE, BootEA, RSN and IMUSE.

\subsubsection{Entity typing.}
Entity typing seeks to predict the ``type entities'' of an instance entity.
It can be regarded as a special link prediction task across an instance KG and an ontological KG. \modelname currently supports TransE-ET, HolE-ET and RESCAL-ET.
Please refer to \cite{FB15K-ET} for more details.

\subsection{Knowledge Graph Embedding Libraries}
As summarized in Table~\ref{tab:summary},
most of existing libraries for KG embeddings only focus on link prediction, a common KG embedding task.
Multi-source KG embedding and entity alignment are only supported by OpenEA \cite{OpenEA} and EAkit \cite{EAkit}.
Only OpenKE \cite{OpenKE}, OpenEA and Pykg2vec \cite{Pykg2vec} are developed with TensorFlow, other libraries only support PyTorch.
LibKGE \cite{LibKGE} is a recent library for link prediction with a high degree of modularity.
DGL-KGE \cite{DGLKE} is developed based on DGL.
It supports PyTorch and XMNet, but not TensorFlow.
NeuralKG \cite{NeuralKG} is a recent Python-based library for diverse representation learning of KGs, but it mainly focuses on rule-based link prediction models.
By contrast, our library is more comprehensive than existing work.
\section{Conclusion and Future Work}

In this paper, we present a new scalable library, \modelname, for multi-source KG embeddings and applications.
It facilitates joint representation learning across multi-source KGs.
It supports PyTorch and TensorFlow2, and can perform multiple tasks, including link prediction, entity alignment, entity typing, and multi-source link prediction, with advanced implementations of the corresponding embedding models.
Extensive experiments validate the effectiveness and efficiency of \modelname.
We further demonstrate how jointly learned embeddings can greatly aid KG-powered downstream tasks such as multi-hop KGQA.
We show that knowledge transfer in multi-source KGs is an efficient way to improve the performance of KG-powered tasks.

\subsubsection{Best practices of KG embedding libraries.}
The proposed \modelname supports multiple tasks, while few libraries support entity typing and multi-source link prediction. 
For users who want to carry out these two tasks, \modelname is the best choice. 
\modelname provides many popular methods in both TensorFlow and PyTorch implementations. 
If the official code of a model only has one implementation but users need another, \modelname is a good choice. 
\modelname is still in its early stages, and a few methods do not achieve optimal results. 
In this case, the original works are more suitable. 
For the models that \modelname currently does not implement, users can try other libraries, e.g., LibKGE \cite{LibKGE} and PyKEEN \cite{PyKEEN} for link prediction, or OpenEA \cite{OpenEA} and EAkit \cite{EAkit} for entity alignment.

\subsubsection{Future work.}
We plan to integrate more KG embedding models and multi-source KG tasks.
We also plan to continually improve our implementations.

\subsubsection{Acknowledgments.} 
This work was supported by National Natural Science Foundation of China (No. 61872172), 
Beijing Academy of Artificial Intelligence (BAAI),
and Collaborative Innovation Center of Novel Software Technology \& Industrialization. 
Zequn Sun was also grateful for the support of Program A for Outstanding PhD Candidates of Nanjing University. 

\bibliographystyle{splncs04}
\bibliography{reference}

\end{document}